\documentclass[11pt]{article}
\usepackage{acl2017}
\usepackage{times}
\usepackage{url}
\usepackage{latexsym}
\usepackage{xspace}
\usepackage{booktabs}
\usepackage{color}
\usepackage{graphicx}
\usepackage{tabulary}
\usepackage{enumitem}
\usepackage{subcaption}
\usepackage{lipsum}

\usepackage{caption}
\usepackage{subcaption}

\usepackage{amsfonts}

  {\begin{itemize}[topsep=0pt, partopsep=0pt] %
    \setlength{\itemsep}{0pt}%
    \setlength{\parskip}{0pt}%
    }%
  {\end{itemize}}
  \definecolor{orange}{RGB}{255,127,0}
   \definecolor{Green}{RGB}{0,100,0}
\definecolor{Purple}{RGB}{147,112,219}

  \usepackage{color}
  {\begin{enumerate}≈ \footnotesize%
    \setlength{\itemsep}{0pt}%
    \setlength{\parskip}{0pt}%
    }%
  {\end{enumerate}}

\usepackage{microtype}
\usepackage{multirow}
\usepackage{verbatim}
\usepackage{amsmath,amsthm,amssymb}
\usepackage{array}
\usepackage[scaled=0.86]{helvet}
\usepackage{ifthen}
\usepackage{courier}
\usepackage[linesnumbered,vlined,ruled]{algorithm2e}

\newcommand{\captionfonts}{\small}
\makeatletter  % Allow the use of @ in command names
\long\def\@makecaption#1#2{%
  \vskip\abovecaptionskip
  \sbox\@tempboxa{{\captionfonts #1: #2}}%
  \ifdim \wd\@tempboxa >\hsize
    {\captionfonts #1: #2\par}
  \else
    \hbox to\hsize{\hfil\box\@tempboxa\hfil}%
  \fi
  \vskip\belowcaptionskip}
\makeatother   % Cancel the effect of \makeatletter

\belowdisplayskip \abovedisplayskip

\newcommand{\sts}{{{\textsc{Seq2Seq}}}\xspace}

\title{Neural Net Models of Open-domain Discourse Coherence}
\author{Jiwei Li and Dan Jurafsky \\
Computer Science Department\\
Stanford University, Stanford, USA \\
{\tt jiweil,jurafsky@stanford.edu} \\
}
\date{}
\aclfinalcopy
\begin{document}
\maketitle

\begin{abstract}
Discourse coherence is strongly associated with text quality,
making it important to natural language generation and understanding.
Yet existing models of coherence focus on measuring individual aspects of coherence
(lexical overlap, rhetorical structure, entity centering) in narrow domains.

In this paper, we describe domain-independent neural models
of discourse coherence that are capable of measuring multiple aspects of coherence 
in existing sentences and can maintain coherence while generating new sentences.
We study both
discriminative models that learn to distinguish coherent from incoherent discourse,
and generative models that produce coherent text,
including a novel neural latent-variable Markovian generative model that 
captures the latent discourse dependencies between sentences in a text.

Our work achieves state-of-the-art performance on multiple coherence evaluations,
and marks an initial step in generating coherent texts given discourse contexts. 
\end{abstract}
\section{Introduction}
%Modeling text coherence is essential in generating readable texts and avoiding confusion. 
Modeling discourse coherence (the way parts of a text are linked into a coherent whole) is essential for 
summarization \cite{barzilay2005sentence}, text planning \cite{hovy1988planning,marcu1997local} question-answering \cite{verberne2007evaluating}, and 
even psychiatric diagnosis \cite{elvevaag07,bedi15}.
%, or compression \cite{sporleder2005discourse}. 
%Varieties of tasks can benefit from a practical discourse coherence evaluation system such as summarization \cite{barzilay1999using,barzilay1999information}, text planning \cite{hovy1988planning,marcu1997local}, question-answering \cite{verberne2007evaluating}, or sentence compression \cite{sporleder2005discourse}. 

Various frameworks exist, each tackling aspects of coherence.
%A variety of models or frameworks have been proposed to tackle different aspects of discourse coherence: 
Lexical cohesion \cite{hallidayhasan,morrishirst91} models  chains of words and synonyms. 
Psychological models of discourse 
\cite{foltz1998measurement,foltz2007discourse,mcnamara} use LSA embeddings to generalize lexical cohesion.
Relational models like
RST  \cite{mann1988rhetorical,lascarides1991discourse} 
define relations  that hierarchically structure texts.
%focusing on connectives and other discourse cues.
%Earlier theoretical attempts such as {\it Rhetorical Structure Theory (RST)} \cite{mann1988rhetorical} \cite{lascarides1991discourse} define relations  that lead to hierarchical structure of texts, with a focus on interdependencies between clauses characterized by connectives and causal relationship signals.
% the hierarchical structure for texts
%Literature from psychology \cite{foltz1998measurement,foltz2007discourse} uses LSA sentence representations, focusing on term-term overlap in coherent contexts. 
%Psychologocail Models
The 
entity grid model \cite{barzilay2008modeling}
and its extensions\footnote{Adding coreference \cite{elsner2008coreference}, named entities \cite{eisner2011extending}, discourse relations \cite{lin2011automatically} and 
entity graphs \cite{guinaudeau2013graph}.} 
capture the referential coherence of
entities moving in and out of focus across a text.
Each captures only a single aspect of coherence,
and all focus on scoring existing sentences,
rather than on generating coherent discourse
for tasks like abstractive summarization.

%In this paper, we propose a framework based on neural nets in an attempt to tackle all the aforementioned aspects involved in discourse coherence (e.g., semantic, syntactic, logic, causality, etc) in an unified framework. 
Here we introduce two classes of neural models for discourse coherence.
Our discriminative models induce coherence  
by treating human generated texts as coherent examples and texts with random sentence replacements as negative examples, feeding LSTM sentence embeddings of pairs of
consecutive sentences to a classifier. 
These achieve state-of-the-art ($96\%$ accuracy) 
on the standard domain-specific sentence-pair-ordering dataset \cite{barzilay2008modeling},
but suffer in a larger open-domain setting due to the small semantic space that 
negative sampling is able to cover. 

Our generative models are based on augumenting encoder-decoder models
with latent variables to model discourse relationships across sentences,
including (1) a model that incorporates an HMM-LDA topic model into the generative model 
and (2) an end-to-end model that introduces a Markov-structured neural
latent variable, inspired by recent work on training latent-variable recurrent nets
\cite{bowman2015generating,serban2016hierarchical}.
These generative models obtain the best result on a large open-domain setting,
including on the difficult task of reconstructing the order of
every sentence in a paragraph, and our latent variable generative model
significantly improves the coherence of text generated by the model.

Our work marks an initial step in 
building end-to-end systems to 
evaluate open-domain discourse coherence, 
and more importantly, generating coherent texts given discourse contexts.

%The rest of this paper is organized as follows:
%we describe the discriminative model in Section 2 and different generative models in Section 3. Experimental results are should in Section 4, followed by a brief conclusion. 
%
%we review frameworks widely adopted in analyzing discourse coherence in Section 2. We describe in detail the proposed discriminative model and 
%the generative model in Section 3. Experimental results are shown in Section 4, followed by a brief conclusion in Section 5. 

%In the next section we summarize related work. Section 3 describes the proposed models. Experimental results are illustrated in Section 4, followed by 
%a brief conclusion. 

\section{The Discriminative Model}
The discriminative model treats cliques (sets of sentences surrounding a center sentence) taken from the original articles as coherent positive examples and cliques with random replacements of the center sentence  as negative examples.
The discriminative model can be viewed as an extended version of Li and Hovy's \shortcite{li2014model} model but is practical at large scale\footnote{Li and Hovy's \shortcite{li2014model} recursive neural model operates on parse trees, which does not support batched computation and is therefore hard to scale up.}. We thus make this section  succinct. 
\paragraph{Notations}
Let $C$ denote a sequence of coherent texts taken from original articles generated by humans. 
$C$ is comprised of a sequence of sentences $C=\{{s_{n-L},...,s_{n-1},s_{n}, s_{n+1},...,s_{n+L}}\}$ where
L denotes the half size of the context window. 
 Suppose each sentence $s_n$ consists of a sequence of words $w_{n1},...,w_{nt},...,w_{nM}$, where $M$ is the number of tokens in $s_n$.
Each word $w$ is associated with a $K$ dimensional vector $h_w$ and each sentence is associated with a $K$ dimensional vector $x_s$. 

Each $C$ contains $2L+1$ sentences, and 
 is associated with a $(2L+1)\times K$ dimensional vector obtained by concatenating the representations of its constituent sentences. The
sentence representation is obtained from LSTMs. 
%Due to space limits, please refer to Appendix section for the detail of LSTM models. 
%(See appendix for LSTM model details.)
%For word compositions, we use the representation outputted from the final time step to represent the entire sentence.  
After word compositions, we use the representation output from the final time step to represent the entire sentence.
Another neural network model 
with a {\em sigmoid} function on the very top layer
is employed to map the concatenation of representations of its constituent sentences to a scalar, indicating the probability of the current clique being a coherent one or an incoherent one. 
\paragraph{Weakness}
Two problems  with the discriminative model stand out:
First, it relies on negative sampling to generate negative examples.  
Since the sentence-level semantic space in the open-domain setting is huge, the sampled instances can  only cover a tiny proportion of the possible
negative candidates, and therefore don't cover the space of possible meanings. As we will show in the experiments section, the discriminative model performs competitively in specific domains, but not in the open domain setting. 
Secondly and more importantly, 
 discriminative models are only able to tell whether an already-given chunk of text is coherent or not. 
While they can thus be used in tasks like extractive summarization for sentence re-ordering, they cannot
be used for coherent text generation in tasks like
dialogue generation or abstractive text summarization.

\section{The Generative Model}
We therefore introduce three neural generative models of discourse coherence.
\subsection{Model 1: the \sts Model and its Variations}
In a coherent context, a machine  should be able to guess the next utterance given the preceding ones. 
A straightforward way to do that is to train a \sts model to predict a sentence given its contexts \cite{sutskever2014sequence}. 
 Generating 
sentences based on neighboring sentences resembles 
 skip-thought models  \cite{kiros2015skip}, which build an encoder-decoder model by predicting tokens in neighboring sentences.

\begin{figure}
    \centering
        \includegraphics[width=2.5in]{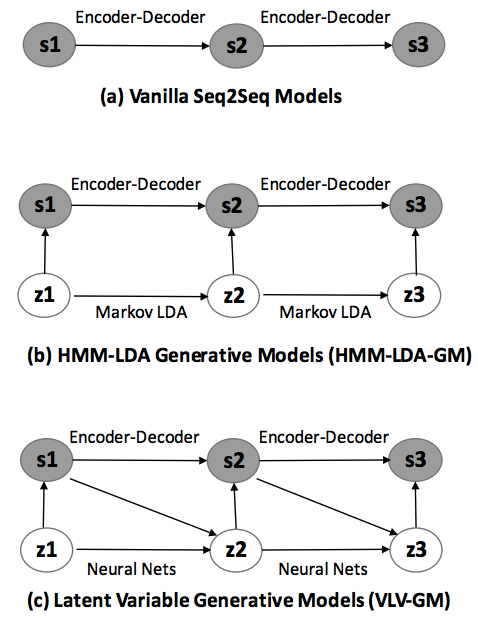}
        \caption{Overview of the proposed generative models for discourse coherence modeling.}
        \label{fig-generative}
\end{figure}

As shown in Figure \ref{fig-generative}a, 
given two consecutive sentences $[s_i, s_{i+1}]$, 
one can measure the coherence by the likelihood of generating $ s_{i+1}$ given its preceding sentence $s_i$ (denoted by {\it uni}).
This likelihood is scaled by the number of words in $s_{i+1}$ (denoted by $N_{i+1}$) to avoid favoring short sequences. 
\begin{equation}
L(s_i,s_{i+1})=\frac{1}{N_{i+1}} \log p(s_{i+1}|s_{i})
\label{eq-mle}
\end{equation}
The probability can be directly computed using a pretrained \sts model \cite{sutskever2014sequence} or an attention-based model \cite{bahdanau2014neural,luong2015effective}. 

In a coherent context, a machine should not only be able to guess the next utterance given the preceding ones, but also the preceding one given the following ones. This gives rise to the 
coherence model (denoted by {\it bi}) that measures the bidirectional dependency between the two consecutive sentences:
\begin{equation}
\begin{aligned}
&L(s_i,s_{i+1})=\frac{1}{N_i}\log p_B(s_i|s_{i+1})\\
&~~~~~~~~~~~~~~~~~~~~~~~~~~+\log \frac{1}{N_{i+1}}p_F(s_{i+1}|s_i)
\end{aligned}
\label{eq-bi}
\end{equation}
%Equ.\ref{eq4} measures the mutual dependency between the two consecutive sentences. 
We separately train two models: a forward model $p_F(s_{i+1}|s_{i})$ 
that predicts the next sentence based on the previous one   
%and $p(s_{i}|s_{i+1})$ that predict the previous sentence givens the next sentence. 
and
a backward model
 $p_B(s_{i}|s_{i+1})$ that predicts the previous sentence given the next sentence. 
$p_B(s_{i}|s_{i+1})$ can be trained in a way similar to $p_F(s_{i+1}|s_{i})$ with sources and targets swapped. It is worth noting that $p_B$ and $p_F$ are separate models and do not share parameters. 

One problem with the described {\it uni} and {\it bi} models is that sentences with higher language model probability (e.g., sentences without rare words) also tend to have higher conditional probability given their preceding or succeeding sentences.  We are interested in 
measuring
the informational gain  from the contexts
 rather than how fluent the current sentence is. 
We thus propose eliminating the influence of the language model, which yields the following coherence score:
\begin{equation}
\begin{aligned}
&L(s_i,s_{i+1})\\
&~~~~~~~~~~=\frac{1}{N_i}[\log p_B(s_i|s_{i+1})-\log p_L(s_i)]\\
&~~~~~~~~~~+\frac{1}{N_{i+1}}[\log p_B(s_{i+1}|s_{i})-\log p_L(s_{i+1})]\\
\end{aligned}
\label{eq-mmi}
\end{equation}
where $p_L(s)$ is the language model probability for generating sentence $s$. We train an LSTM language model, which can be thought of as a \sts model with an empty source. A closer look at Eq. \ref{eq-mmi} shows that it is of the same form as the mutual information between $s_{i+1}$ and $s_i$, namely $\log [p(s_{i+1},s_i)/p(s_{i+1})p(s_i)]$.
\paragraph{Generation}
The scoring functions in Eqs. \ref{eq-mle}, \ref{eq-bi}, and \ref{eq-mmi} are discriminative, generating coherence scores for an already-given chunk of text. 
Eqs. \ref{eq-bi} and \ref{eq-mmi} can not be directly used for generation purposes, 
since they requires the completion of $s_{i+1}$ before the score can be computed.
A normal strategy is to generate a big N-best list using Eq. \ref{eq-mle} and then rerank the 
N-best list using Eq. \ref{eq-bi} or \ref{eq-mmi}
\cite{li2015diversity}. 
The N-best list can be generated using standard beam search, or other algorithmic variations that promote diversity, coherence, etc. \cite{shao2017generating}. 
\paragraph{Weakness}
(1) The \sts model generates words sequentially based on an evolving hidden vector, 
which is updated by combining the current word representation with
previously built hidden vectors.   
The generation process is thus not exposed to more global features of the discourse
like topics. 
As the hidden vector evolves, the influence from contexts  gradually diminishes, with language models quickly dominating.
(2) By predicting a sentence conditioning only on its left or right neighbor, the model lacks the ability to  handle the longer-term  discourse dependencies across
the sentences of a text.

To tackle these two issues, we need a model that is able to constantly remind the decoder 
about the global meaning that it should convey
at each word-generation step, a global meaning which can capture the
state of the discourse  across the sentences of a text.
We  propose two models of this global meaning,
a pipelined approach 
based on HMM-based topic models \cite{blei2003latent,gruber2007hidden},
and an end-to-end generative model with variational latent variables.

\subsection{HMM-LDA based Generative Models (HMM-LDA-GM)}
In Markov topic models the topic depends 
on the previous topics in context
 \cite{ritter2010unsupervised,paul2010two,wang2011structural,gruber2007hidden,paul2012mixed}.
The topic for the current sentence is drawn
based on the topic of the preceding sentence (or word) 
rather than on the global document-level topic distribution in vanilla LDA.

Our first model is a pipelined one
(the {\it HMM-LDA-GM} in  Fig. \ref{fig-generative}b),
in which
an HMM-LDA model provides the \sts model with global information for token generation,
with two components:
 
 (1) {\bf Running HMM-LDA}:  we first run a sentence-level HMM-LDA similar to 
\newcite{gruber2007hidden}.
Our implementation forces
 all words in a sentence to be generated from the same topic, and this topic is sampled from a distribution based on the topic from previous sentence. 
  Let $t_n$ denote the distribution of topics for the current sentence, where $t_n\in \mathbb{R}^{1\times T}$.
We also associate  
each LDA topic with a $K$ dimensional vector, representing the semantics embedded in this topic. The topic-representation matrix is denoted by $V\in  \mathbb{R}^{T\times K}$, where $T$ is the pre-specified number of topics in LDA. 
$V$ is learned in the word predicting process when training  encoder-decoder models. 

(2) {\bf Training encoder-decoder models}: 
For the current sentence $s_n$, given its topic distribution $t_n$, we first compute the topic representation $z_n$ for $s_n$ using the weighted sum of LDA topic vectors:
\begin{equation}
z_n=t_n\times V
\end{equation} 
$z_n$ can be thought of as a discourse state vector that stores the information the current sentence needs to convey in the discourse, and is 
used to guide 
 every step of word generation in $s_n$. 
We run the encoder-decoder model, which subsequently predicts tokens in $s_n$ given $s_{n-1}$. This process is the same as the vanilla version of \sts models, the only difference being that $z_n$ is incorporated into each step of decoding for hidden vector updates:
\begin{equation}
p(s_n|z_n,s_{n-1})=\prod_{t=1}^M p(w_t|h_{t-1},z_n)
\label{eq-generative}
\end{equation}  
 $V$ is updated along with parameters in the encoder-decoder model. 

$z_n$ influences each time step of decoding, and thus  
addresses the problem that vanilla \sts models gradually lose
global information as the hidden representations evolve.
$z_n$ is computed based on the topic distribution $t_n$, which is obtained from the HMM-LDA model, thus modeling the global Markov discourse dependency between sentences of the text.\footnote{This pipelined approach is closely related to recent work that incorporates LDA topic information into generation models in an attempt to leverage context information \cite{ghosh2016contextual,xing2016topic,mei2016coherent}} 
The model can be adapted to the bi-directional setting, in which we separately train two models to handle the forward probability $\log p(t_n|s_{n-1},...)$ and the backward one $\log p(t_n|s_{n+1})$. 
The bi-directional ({\it bi}) strategy described in Eq.~\ref{eq-mmi} can also be incorporated to remove the influence of language models.

\paragraph{Weakness} Topic models (either vanilla or HMM versions) focus on word co-occurrences  at the document-level and are thus very lexicon-based. 
Furthermore, given 
the diversity of topics in a dataset like Wikipedia 
but the small number of topic clusters,
the LDA model usually produces very coarse-grained topics 
(politics, sports, history, etc.), assigning
very similar topic distributions to consecutive sentences.
These topics thus capture topical coherence
but are too coarse-grained to capture all the more fine-grained aspects of
{\em discourse} coherence relationships.

\subsection{Variational Latent Variable Generative Models (VLV-GM)}
We therefore propose instead to
train an end-to-end system, in which the meaning transitions between  sentences can be naturally learned from the data. 
Inspired by recent work on generating sentences from a latent space \cite{serban2016hierarchical,bowman2015generating,chung2015recurrent},
we propose the VSV-GM model in Fig. \ref{fig-generative}c.
Each sentence $s_n$ is again associated with a hidden vector representation $z_n\in  \mathbb{R}^K$ which stores the global information that the current sentence needs to talk about, but instead of obtaining $z_n$ from an upstream model like LDA, $z_n$ is learned from the training data. 
$z_n$ is a stochastic latent variable conditioned
on all previous sentences and $z_{n-1}$: 
\begin{equation}
\begin{aligned}
&p(h_n|h_{n-1},s_{n-1},s_{n-2},...)=N(\mu^{\text{true}}_{h_n},\Sigma^{\text{true}}_{h_n})\\
&\mu^{\text{true}}_{h_n}=f(h_{n-1},s_{n-1}, s_{n-2}, ...)\\
&\Sigma^{\text{true}}_{h_n}=g(h_{n-1},s_{n-1}, s_{n-2}, ...)\\
\end{aligned}
\label{eq-p}
\end{equation}

where $N (\mu,\Sigma)$ is a multivariate normal distribution with mean $\mu\in  \mathbb{R}^K$
 and covariance matrix
 $\Sigma\in  \mathbb{R}^{K\times K}$. $\Sigma$ is a diagonal matrix. 
As can be seen, the global information $z_n$ for the current sentence depends on the information $z_{n-1}$ for its previous sentence as well as the 
text of the  context sentences.
This forms a Markov chain across all sentences. 
$f$ and $g$ are neural network models that take previous sentences
and $z_{n-1}$,
 and map them to a real-valued representation using hierarchical LSTMs \cite{li2015hierarchical}\footnote{Sentences are first mapped to vector representations using a LSTM model. Another level of LSTM at the sentence level then composes representations of the multiple sentences to a single vector.}.

Each word $w_{nt}$ from $s_n$ is predicted using the concatenation of the representation previously build by the LSTMs (the same vector used in word prediction in vanilla \sts models) and $z_n$, as shown in Eq.\ref{eq-generative}. 

We are interested in the posterior distribution $p(z_n|s_1,s_2,...,s_{n-1})$, namely, the information that the current sentence needs to convey given the preceding ones. 
Unfortunately,  a highly non-linear mapping from $z_n$ to tokens in $s_n$ 
results in intractable inference of the
posterior. A common solution is to use variational inference to learn another distribution, denoted by $q(z_n|s_1,s_2,...,s_N)$, to approximate the true posterior $p(z_n|s_1,s_2,...,s_{n-1})$. 
The model's latent variables are obtained by maximizing the variational lower-bound of observing the dataset:
\begin{equation}
\begin{aligned}
&\log p(s_1,..,s_N)\leq\\
&\sum_{t=1}^N-D_{KL}(q(z_n|s_n,s_{n-1},...)||p(z_n|s_{n-1},s_{n-2},...))\\
&+E_{q(z_n|s_n,s_{n-1},...)}\log p(s_n|z_n,s_{n-1},s_{n-2},...)
\end{aligned}
\label{lower-bound}
\end{equation}
This objective to optimize consists of two parts; the first is the KL
divergence  
between the approximate distribution $q$ and 
  the true posterior $p(s_n|z_n,s_{n-1},s_{n-2},...)$, in which we want to approximate the true posterior using $q$. 
The second part $E_{q(z_n|s_n,s_{n-1},...)}\log p(s_n|z_n,s_{n-1},s_{n-2},...)$,
predicts tokens in $s_n$ in the same way as in \sts models 
with the difference that it considers the global information $z_n$.

The approximate posterior distribution $q(z_n|s_n,s_{n-1},...)$ takes a form similar to $p(z_n|s_{n-1},s_{n-2},...)$:
\begin{equation}
\begin{aligned}
&q(z_n|s_{n},s_{n-1},...)=N(\mu^{\text{ approx}}_{z_n},\Sigma^{\text{ approx}}_{z_n})\\
&\mu^{\text{ approx}}_{z_n}=f_q(z_{n-1},s_{n}, s_{n-1}, ...)\\
&\Sigma^{\text{ approx}}_{z_n}=g_q(z_{n-1},s_{n}, s_{n-1}, ...)\\
\end{aligned}
\label{q}
\end{equation}  
$f_q$ and $g_q$ are of similar structures to $f$ and $g$, using a hierarchical neural network model to map context tokens to vector representations. 
\paragraph{Learning and Testing}
At training time, the approximate posterior $q(z_n|z_{n-1},s_{n},s_{n-1},...)$, 
the true distribution $p(z_n|z_{n-1},s_{n-1},s_{n-2},...)$, and the generative probability 
$p(s_n|z_n,s_{n-1},s_{n-2},...)$
 are trained jointly 
 by maximizing the
variational lower bound with respect to their parameters: 
a sample $z_n$ is first drawn from the posterior distribution $q$,
namely $N(\mu^{\text{ approx}}_{z_n},\Sigma^{\text{ approx}}_{z_n})$. 
This sample is used to approximate the expectation  $E_{q}\log p(s_n|z_n,s_{n-1},s_{n-2},...)$. 
Using $z_n$, we can update the encoder-decoder model using SGD in a way similar to the standard \sts model, the only difference being that the current token to predict not only depends on the LSTM output $h_t$, but also $z_n$.
Given the sampled $z_n$, the KL-divergence can be readily computed, and we update the model using standard gradient decent (details shown in the Appendix). 

The proposed {\it VLV-GM} model can  be adapted to the bi-directional setting and the {\it bi} setting similarly to the way  LDA-based models are adapted. 

The proposed model is closely related to many recent attempts 
 in training variational autoencoders (VAE) \cite{kingma2013auto,rezende2014stochastic}, variational or latent-variable recurrent nets \cite{bowman2015generating,chung2015recurrent,ji2016latent,bayer2014learning}, hierarchical latent variable encoder-decoder models \cite{serban2016hierarchical,serban2016multiresolution}.

\section{Experimental Results}

In this section, we describe experimental results. We first evaluate the proposed models on 
discriminative tasks such as sentence-pair ordering and full paragraph ordering reconstruction.
Then we look at the task of coherent text generation.  

%In this section, we detail experimental results. 
\begin{table}[htb]
\centering
\small
\begin{tabular}{cccc}\\\hline
Model&Acci&Earthq&Aver\\\hline
Discriminative Model &{\bf 0.930}&{\bf 0.992}&{\bf 0.956}\\
 \sts (bi)&0.755&0.930&0.842\\
VLV-GM (bi)&0.770&0.931&0.851\\
\hline
Recursive & 0.864&0.976&0.920\\
Entity Grid Model&  0.904 &0.872& 0.888\\\hline
HMM &0.822 &0.938&0.880  \\
HMM+Entity &0.842 &0.911&0.876 \\
HMM+Content & 0.742 &0.953&0.847\\
Graph  &0.846 &0.635&0.740 \\\hline
\newcite{foltz1998measurement}-Glove&0.705&0.682 &0.688\\
\newcite{foltz1998measurement}-LDA&0.660&0.667&0.664 \\
\end{tabular}
\caption{
Results from different coherence models. Results for the Recursive  model is reprinted from \newcite{li2014model}, Entity Grid Model from \newcite{louis2012coherence}, HMM, HMM+Entity and HMM+Content from \newcite{louis2012coherence}, Graph from \newcite{guinaudeau2013graph}, and the final two lexical models are recomputed
using Glove and LDA to replace the original LSA model of \newcite{foltz1998measurement}.}
%(Barzilay and Lapata, 2008; Louis and Nenkova, 2012; Guinaudeau and Strube, 2013).}
\label{tab:domspec}
\end{table}
%\subsection{Sentence Ordering on Domain-specific Datasets}
\subsection{Sentence Ordering, Domain-specific Data }
\paragraph{Dataset}

%We first evaluate the proposed coherence models on a  evaluation dataset widely adopted in existing work \cite{barzilay2008modeling,louis2012coherence,elsner2007unified,lin2011automatically} for sentence ordering. 
We first evaluate  the proposed algorithms on the task of predicting
the correct ordering of pairs of sentences 
predicated on the assumption that an article is always
more coherent than a random permutation of its sentences \cite{barzilay2008modeling}.
A detailed description of this commonly used dataset and  
training/testing are found in the Appendix.

We report the performance of the following baselines widely used in the coherence literature.

(1) {\it Entity Grid Model}: The grid model 
presented in \newcite{barzilay2008modeling}. Results are directly taken from Barzilay and Lapata's \shortcite{barzilay2008modeling} paper. We also consider
variations of entity grid models, such as
Louis and Nenkova (2012)  which models the cluster transition probability  and the {\it Graph Based Approach}  which uses
a graph to represent the entity transitions
needed for local coherence computation \cite{guinaudeau2013graph}.

(2) \newcite{li2014model}: A recursive neural model computes sentence representations based on parse trees. Negative sampling is used to construct  negative incoherent examples. Results are from their papers.

(3) 
\newcite{foltz1998measurement} computes the semantic relatedness of two text units as the cosine similarity between their LSA vectors. The coherence of a discourse is
the average of the cosine of adjacent sentences. 
We used this intuition, but with more modern embedding models:
(1) 300-dimensional Glove word vectors \cite{pennington2014glove}, embeddings for a sentence computed by averaging the embeddings of its words (2) Sentence representations obtained from LDA \cite{blei2003latent} with 300 topics, trained on the Wikipedia dataset. 
\begin{comment}
We compute  coherence as the average cosine between adjacent sentences.
\end{comment}
Results are reported in Table 2. 
\begin{comment}
The {\it domain specific} settings, which use an identical training set as the baselines,
perform slightly better than the neural models of Li et al., (2014) and 
outperform all other baselines.  This
confirms the effectiveness of the proposed framework on the algorithmic level. 
When the proposed model is trained on large scale datasets,
it surpasses all existing baselines even without using the training dataset ({\it original}).
The domain adaptation model, which both takes advantage of massive training data  and also adapts to the domain-specific setting of the training set, 
 yields the new  state-of-the-art.
 \end{comment}
 The extended version of the discriminative model described in this work
 significantly outperforms the parse-tree based recursive models presented in \newcite{li2014model} as well as all non-neural baselines. It achieves almost perfect accuracy on the earthquake dataset and $93\%$ on the accident dataset, marking a significant advancement in the benchmark. 
Generative models (both vanilla \sts and the proposed variational model) do not perform competitively on this dataset. 
We conjecture that 
this is due to the small size of the dataset, leading the generative model to overfit.
\begin{comment} 
 %The simple method of calculating cosine similarity between adjacent sentences, adopted from \shortcite{foltz1998measurement} does not yield competitive results, confirming that lexical coherence is only one component of discourse coherence. 
 The simple LSA method of calculating cosine similarity between adjacent sentences,
 adopted from \newcite{foltz1998measurement}, does not yield competitive results,
 confirming that while simple centroids of word embeddings may do a good job of modeling 
 lexical coherence, lexical coherence is only one component of discourse coherence. 
 \end{comment}
  
\subsection{Evaluating Ordering on Open-domain} 

Since the dataset presented in \newcite{barzilay2008modeling} is quite domain-specific, 
%we also built an open-domain dataset of much larger scale. 
we propose testing coherence with a much larger, open-domain dataset: Wikipedia.
%Our test set consists of 984 paragraphs from Wikipedia, each of which is consists 16 sentences.
We created a test set by randomly selecting 984 paragraphs from Wikipedia dump 2014,
each paragraph consisting of at least 16 sentences. The training set is 
30 million sentences not overlapping with the test set.
%We take a subset of the dataset, containing roughly 30 million sentences.

\subsubsection{Binary Permutation Classification}
We adopt the same strategy as in \newcite{barzilay2008modeling}, in which we generate pairs of sentence permutations from the original Wikipedia paragraphs. 
We follow the protocols described in the subsection and 
each pair whose original paragraph's score is higher than its permutation is treated as being correctly classified, else incorrectly classified. 
Models are evaluated using accuracy. 
We  implement the Entity Grid Model \cite{barzilay2008modeling} using the Wikipedia training set as a baseline, the detail of which is presented in the Appendix. 
Other baselines consist of the Glove and
LDA updates of the lexical coherence baselines
(Foltz et al., 1998).

\paragraph{Results}
Table \ref{open-domain} presents results on the binary classification task. 
Contrary to the findings on the domain specific dataset in the previous subsection, 
%we find that the proposed discriminative does not yield compelling results, performing only slightly better than the {\it entity grid model}. 
the discriminative model does not yield compelling results, performing only slightly better than the entity grid model.
%The explanation is as follows: 
%the proposed discriminative model relies on sentence-level negative sampling. 
We believe the poor performance is due to the sentence-level negative sampling used by the discriminative model.
Due to the huge semantic space in the open-domain setting, the sampled instances can only cover a tiny proportion of the possible
negative candidates, and therefore don't cover the space of possible meanings.
By contrast the dataset in \newcite{barzilay2008modeling} is very domain-specific,
and the semantic space is thus relatively small. 
By treating all other sentences in the document as negative, the discriminative strategy's negative samples form
a much larger proportion of the semantic space, leading to good performance. 

Generative models perform significantly better than all other baselines. 
Compared with the dataset in \newcite{barzilay2008modeling},   overfitting is not an issue here due to the great amount of training data.  
In line with our expectation,
the {\it MMI} model outperforms the {\it bidirectional} model, which in turn outperforms the {\it unidirectional} model across all three generative model settings.
We thus only report {\it MMI} results for experiments below.
 The {\it VLV-GM} model outperforms that the {\it LDA-HMM-GM} model, which is slightly better than the vanila \sts models. 

\begin{table}
\small
\centering
\begin{tabular}{cc}\hline
 Model&Accuracy\\\hline
 VLV-GM (MMI)&{\bf 0.873}\\
  VLV-GM (bi)&0.860\\
   VLV-GM (uni)&0.839\\
 \hline
   LDA-HMM-GM   (MMI)&0.847\\
 LDA-HMM-GM  (bi)&0.837\\
 LDA-HMM-GM  (uni)&0.814\\\hline
 \sts   (MMI)&0.840\\
\sts   (bi)&0.821\\
\sts   (uni)&0.803\\\hline
Discriminative Model&0.715\\\hline
Entity Grid Model&0.686 \\
\newcite{foltz1998measurement}-Glove&0.597 \\
\newcite{foltz1998measurement}-LDA&0.575 \\
 \end{tabular}
 \caption{Performance on the open-domain binary classification dataset of 984 Wikipedia paragraphs.}
 \label{open-domain}
 \end{table}

\begin{table}
\centering
\small
\begin{tabular}{lc}\hline
 Model&Accuracy\\\hline
VLV-GM (MMI)&{\bf 0.256}\\
  LDA-HMM-GM (MMI)&0.237 \\
 \sts   (MMI)&0.226\\
Entity Grid Model&0.143\\
\newcite{foltz1998measurement} (Glove)& 0.084\\
 \end{tabular}
 \caption{Performances of the proposed models on the open-domain paragraph reconstruction dataset.}
 \label{fig:reconstruction}
 \end{table}

\subsubsection{Paragraph Reconstruction}

%The accuracy of our models on the binary task of detecting original
The accuracy of our models on the binary task of detecting the original sentence
ordering is very high, on both the prior small task and our large
open-domain version.  We therefore believe it is time for the community
to move to a more difficult task for measuring coherence.

%We use the task of
%{\em reconstructing an original paragraph from a bag of  constituent sentences},
%which has been widely adopted in coherence evaluation \cite{lapata2003probabilistic}. 

We suggest the task of
{\em reconstructing an original paragraph from a bag of  constituent sentences},
which has been previously used in coherence evaluation \cite{lapata2003probabilistic}.
More formally, given a set of  permuted sentences $s_1,s_2, ..., s_N$ (N the number of sentences in the original document), our goal is return the original 
(presumably most coherent) ordering of $s$. 

Because the {\it discriminative model} calculates the coherence of a sentence 
given the known previous and following sentences, it cannot be applied to this task
since we don't know the surrounding context.  Hence, we only use 
the {\it generative model}.  
The first sentence of a paragraph is given: for each step, we compute the coherence score of placing each remaining candidate sentence to the right of the partially constructed document. 
We use beam search with beam size 10.
\begin{comment}
(2)
No clue is given: we employed the graph based method described in \newcite{lapata2003probabilistic}. 
We first construct a graph where the each vertex denotes a sentence and the edge weight $u\rightarrow v$ denotes the coherence score of sentence $v$ coming after $u$. 
Note that  weight values for  $u\rightarrow v$ and $v\rightarrow u$ are different. 
We initialize the vertex list $V$ using all vertexes in the graph. 
Similar to \newcite{lapata2003probabilistic}, we employ a greedy search model. The greedy algorithm first picks the edge $u\rightarrow v$ with the highest coherence score, and deletes all the outgoing edges from vertex $u$ and all incoming edges to vertex $v$. 
$u, v$ are removed from the vertex list $V$. 
Next, for each time step, let $v_{\text{left}}$ and $v_{\text{right}}$ respectively denote the left-most and right-most node in the partially constructed paragraph. 
The greedy model chooses whether to expand the paragraph to the left and to the right by comparing $\text{max}_{v'\in V} \text{S}(v', v_{\text{left}})$ with $\text{max}_{v'\in V} \text{S}(v_{\text{right}}, v')$, where the former denotes the maximal coherence score of placing a remaining sentence to the left of the partially constructed paragraph and the latter denotes the maximal score of appending a sentence to the right of the paragraph. 
The newly selected node is added to the paragraph and removed from the vertex list $V$. We repeat this process until $V$ is empty. 
\end{comment}
We use the Entity Grid model as a baseline for both the settings. 

Evaluating the absolute positions of sentences would be too harsh,
penalizing orderings that  maintain relative position between
sentences through which local coherence can be manifested. We therefore
use
  Kendall's $\tau$ \cite{lapata2003probabilistic,lapata2006automatic}, a metric of rank correlation for evaluation. See the Appendix for details of Kendall's $\tau$ computation.
 We observe a pattern similar to the results on the binary classification task, where the {\it VLV-GM} model performs the best. 
\begin{table}
\small
\centering
\begin{tabular}{cccc}\hline
Model& adver-1 &adver-2  &adver-3    \\\hline
VLV-GM (MMI) &{\bf 0.174}&{\bf 0.120}&{\bf 0.054} \\
LDA-HMM-GM  (MMI)&0.130&0.104&0.043\\
 \sts (MMI)&0.120  &0.090 &0.039  \\
   \sts (bi)&0.108&0.078&0.030 \\
 \sts (uni)&  0.101 & 0.068& 0.024 \\
\end{tabular}
\caption{Adversarial Success for different models.}
\label{adver}
\end{table}

\subsection{Adversarial evaluation on Text Generation Quality}
Both the tasks above are discriminative ones. We also want to evaluate different models' ability to generate coherent text chunks. The experiment is set up as follow: each encoder-decoder model is first given a set of context sentences (3 sentences). The model 
then
 generates a succeeding sentence   
 using beam-search given the contexts. 
For the {\it uni-directional} setting, we directly take the most probable sequence 
and for the {\it bi-directional} and {\it MMI}, we rerank the N-best list using the backward probability and language model probability. 

We conduct experiments on multi-sentence generation, in which we repeat the generative process described above for $N$ times, where $N$=1,2,3.  At the end of each turn, the context is updated by adding in the newly generated sequence, and this sequence is used as the source input to the encoder-decoder model for next sequence generation. 
For example, when $N$ is set to 2, given the three context sentences {\it context-a}, {\it context-b} and {\it context-c}, we first generate {\it sen-d} given the three context sentences and then generate {\it sen-e} given the {sen-d},  {\it context-a}, {\it context-b} and {\it context-c}. 

For evaluation, standard word overlap metrics such as BLEU or ROUGE are not suited for our task, and we use adversarial evaluation \newcite{bowman2015generating,kannan}.
In adversarial evaluation, we
train a binary  discriminant function to classify
a sequence as machine generated or human generated,
in an attempt to evaluate the model's sentence generation capability.
The evaluator takes as input the concatenation of the contexts and the generated sentences (i.e.,
 {\it context-a}, {\it context-b} and {\it context-c}, {\it sen-d} , {\it sen-e} in the example described above),\footnote{The model uses a hierarchical neural structure that first maps each sentence to a vector representation, with another level of LSTM on top of the constituent sentences, producing a single vector to represent the entire chunk of texts. 
} and outputs a scalar, indicating the probability of the current text chunk being human-generated. Training/dev/test sets are held-out sets from the one on which generative models are trained. They respectively contain 128,000/12,800/12,800 instances.
Since discriminative models cannot generate sentences, and thus cannot be used for adversarial evaluation, they are skipped in this section. 
  
We report Adversarial Success ({\it AdverSuc} for short), which is the fraction of instances in which a model is capable of fooling the evaluator. {\it AdverSuc} is the difference between 1 and the accuracy 
achieved by the  evaluator. Higher values of {\it AdverSuc} for a dialogue generation model are better. 
{\it AdverSuc-N} denotes the adversarial accuracy value on machine-generated texts with N turns. 

Table \ref{adver} show 
{\it AdverSuc} numbers for different models.
As can be seen, the latent variable model {VLV-GM} is able to generate chunk of texts that are most indistinguishable from coherent texts from humans. This is due to its ability to handle the dependency between neighboring sentences. 
Performance declines as the number of turns increases due to the accumulation of errors and 
current
 models' inability to model long-term sentence-level dependency. 
All  models perform poorly on the {\it adver-3} evaluation metric, with the best adversarial success value being 0.081 (the trained evaluator is able to  distinguish between human-generated and machine generated dialogues with greater than 90 percent accuracy for all models).
\begin{figure}
\hspace{-0.5cm}
        \includegraphics[width=3.3in]{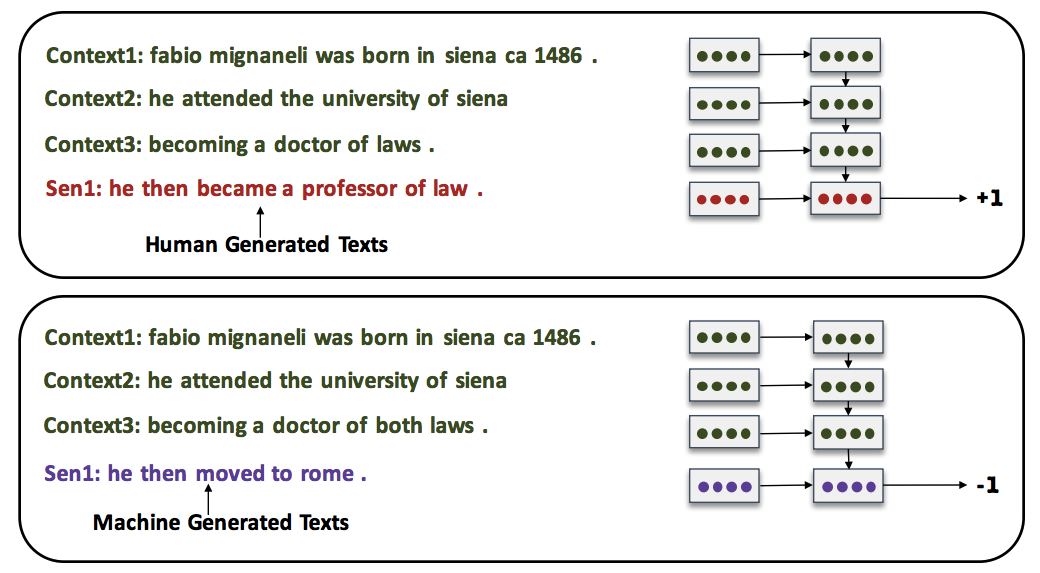}
        \caption{An overview of training the adversarial evaluator using a hierarchical neural model. {\color{Green} Green} denotes input contexts. {\color{red} Red} denotes a sentence from human-generated texts, treated as a positive example. {\color{Purple} Purple} denotes a sentence from machine-decoded texts, treated as a negative example.}
        \label{adverfig}
\end{figure}

\subsection{Qualitative Analysis}

With the aim of guiding future investigations,
we also briefly explore our model qualitatively,
examining the coherence scores assigned to some 
artificial miniature discourses that exhibit various kinds of
coherence.

\paragraph{Case 1: Lexical Coherence} ~\\
\begin{small}
\noindent
{\it Pinochet was arrested.  His arrest was unexpected.} {\bf 1.79}\\
{\it Pinochet was arrested. His death was unexpected. }  {\bf 0.84}\\
{\it Mary ate some apples. She likes apples.} {\bf 2.03}\\
{\it Mary ate some apples. She likes pears.} {\bf  0.27}\\
{\it Mary ate some apples. She likes Paris.} {\bf -1.35}\\

\end{small}

The examples suggest that the model handles lexical coherence,
correctly favoring the 1st over the 2nd, and the 3rd over the 4th examples.
Note that the coherence score for the final example is negative,
which means conditioning on the first sentence actually decreases the likelihood of generating the second one. 

%By comparing example 1 with 2, 3 with 4, we can see that the model favors cliques with higher percentage of lexical overlaps. The model favors  the third example over the fourth, but still tends to think the fourth example as coherent by assigning its likelihood of being coherent more than 0.5. 
%This is because word ``apples" and ``pears" are usually close in the vector space. 
%When the model composes constituent words into sentence-level embeddings, the two words would not contribute significantly 
%to the sentence embedding that will lead the model to make opposite decisions.
% How to deal with these issues constitute our future work. 

\paragraph{Case 2: Temporal Order} ~\\

\begin{small} 

\vspace*{-12pt}
\noindent
{\it Washington was unanimously elected president in the first two national elections. He oversaw the creation of a strong, well-financed national government.} {\bf  1.48} \\
%{\it 1: Washington was unanimously elected president in the first two national elections. He oversaw the creation of a strong, well-financed national government.} P(coherent)=0.782. \\
{\it Washington oversaw the creation of a strong, well-financed national government. He was unanimously elected president in the first two national elections.} {\bf 0.72}
% P(coherent)=0.324.\\

\end{small} 
 
\paragraph{Case 3: Causal Relationship} ~\\

\begin{small} 

\vspace*{-12pt}
\noindent
{\it Bret enjoys video games; therefore, he sometimes is late to appointments. } {\bf 0.69}  ~\\
{\it Bret sometimes is late to appointments; therefore, he enjoys video games. }  {\bf -0.07}\\

\end{small} 

Cases 2 and 3 suggest the model may, at least in these simple cases, be capable of addressing
the much more complex task of dealing with temporal
and causal relationships.  Presumably this is because
the model is exposed in training to the general preference of natural text for
temporal order, and even for the more subtle causal links.

\paragraph{Case 4: Centering/Referential Coherence} ~\\

\begin{small} 

\vspace*{-12pt}
 \noindent{\it Mary ate some apples. She likes apples.} {\bf 3.06}\\
 \noindent{\it She ate some apples. Mary likes apples.} {\bf 2.41}\\

\end{small} 
 
The model seems to deal with simple cases of referential coherence.
\medskip

\begin{small} 

\noindent
{\it Example3: {\bf 2.40}\\
 John went to his favorite music store to buy a piano.  ~\\
 He had frequented the store for many years. \\
 He was excited that he could finally buy a piano. \\
  He arrived just as the store was closing for the day. 
 }
 \\
 {\it Example4: {\bf 1.62}\\
 John went to his favorite music store to buy a piano.  ~\\
It was a store John had frequented for many years ~\\
He was excited that he could finally buy a piano.. \\
  It was closing just as John arrived.
 }\\

\end{small} 

In these final examples from \newcite{miltsakaki2004evaluation},
the model successfully captures the fact that
the second text is less coherent due to  {\em rough shifts}.
This suggests that the discourse embedding space may be able to capture
a representation of entity focus.

Of course all of these these qualitative evaluations are only suggestive,
and a deeper understanding of what the discourse embedding space
is capturing will likely require more sophisticated visualizations.

\begin{comment}
\paragraph{Case3: Coreference} ~\\
{\it Example1: Mary likes some apples. She like.} P(coherent)=0.742.  ~\\
{\it Example2: Mary likes read apples. But he does not like red ones. }  P(coherent)=0.684 ~~\\
Apparently, the second example misuses the third-person singular pronoun ``he". However, the model is still quite positive about its being coherent. The explanation about why the models fails to make correct decision here are three folds:  
(1) During training, negative examples are randomly selected. The chance that we select sentences whose the overall meaning of sentence is similar to the original one but one particular pronoun is misused is very small. Cases similar to the second example  therefore rarely appear as negative in the training set. (2) Words ``she" and ``he" are relatively close in the semantic space. When we compose words into sentence-level embeddings, the two words would not contribute significantly 
to 
 the sentence embedding to lead the model to make opposite decisions. 
(3) It requires perfect the modeling of  long-term dependency to  link third-person singular pronoun  in the second sentence back to ``Mary" in the first sentence. 
%The fact that relatively small progress has made in coreference literature also validates this point. 
One transitional criticism  leveraged against deep learning models is its incapability of modeling long-term dependency \cite{bengio1994learning}. 
Exploring attention models to address this issue naturally constitutes our future work. 
\end{comment}

\section{Conclusion}
We investigate the problem of open-domain discourse coherence,
training discriminative models that
treating natural texts as coherent and permutations as
non-coherent, and Markov
generative models that can predict sentences given their neighbors. 

Our work shows that the traditional evaluation metric (ordering pairs
of sentences in small domains) is completely solvable by our discriminative models,
and we therefore suggest the community move to the harder task
of open-domain full-paragraph sentence ordering. 

The proposed models also offer an initial step in generating coherent texts given contexts,
which has the potential to benefit a wide range of 
generation tasks in NLP. Our latent variable neural models, by 
offering a new way to learn latent discourse-level features of a text,
also suggest new directions in discourse representation
that may bring benefits to any discourse-aware NLP task.

 \paragraph{Acknowledgements}
The authors thank Will Monroe, Sida Wang, Kelvin Guu and the other members of the Stanford NLP Group
for helpful discussions and comments. 
Jiwei Li is supported by a Facebook Fellowship,  which we gratefully acknowledge. 
This work is also partially supported by 
the NSF under award IIS-1514268, and
the DARPA Communicating
with Computers (CwC) program under ARO prime
contract no. W911NF- 15-1-0462. Any opinions,
findings, and conclusions or recommendations expressed
in this material are those of the authors and
do not necessarily reflect the views of DARPA, the NSF,
or Facebook.

%However, despite achieving state of the art performance,
%our models sometimes fail to handle the focus shifts necessary  to handle
%centering-style referential coherence.  This suggests future work
%integrating our model with centering-based grid models,
%or perhaps learning a model that latently captures entity focus across a discourse.

%, obviated the need for intensive efforts
%
%involved in feature engineering.  To the best of our knowledge,
%this paper represents the first work for open-domain coherence
%system.

\bibliographystyle{acl_natbib}
\bibliography{coherence}

\section{Supplemental Material}
\label{sec:supplemental}

\paragraph{Details for the domain specific dataset \cite{barzilay2008modeling}}
The corpus consists of 200 articles each from two domains:
NTSB airplane accident reports (V=4758, 10.6 sentences/document)
and AP earthquake reports  (V=3287, 11.5 sentences/document),
split into training and testing.
For each document, pairs of permutations are generated\footnote{Permutations downloaded from
\url{people.csail.mit.edu/regina/coherence/CLsubmission/}.}.  
Each pair contains the original document order and a random permutation
of the sentences from the same document.
\paragraph{Training/Testing details for models on the domain specific dataset }
We use reduced versions of both generative and discriminative models to allow fair comparison with baselines.
For the discriminative model, we generate noise negative examples from random replacements in the training set, with the only difference that random replacements only come from the same document.
We use 300 dimensional embeddings borrowed from GLOVE \cite{pennington2014glove} to initialize word embeddings. 
Word embeddings are kept fixed during training and we update LSTM parameters using AdaGrad \cite{duchi2011adaptive}. 
For the generative model, due to the small size of the dataset, we train a one layer  \sts model with word dimensionality and number of hidden neurons set to 100. 
The model is trained using SGD with AdaGrad \cite{zeiler2012adadelta}.

The task requires a coherence score for the whole document, which is comprised of multiple cliques. We adopt the strategy described in \newcite{li2014model} by breaking the document into a series of cliques which is comprised of a sequence of consecutive sentences. The document-level coherence score is attained by averaging its constituent  cliques. We say a document is more coherent if it achieves a higher average score within its constituent cliques.

\paragraph{Implementation of Entity Grid Model}
For each noun in a sentence, we extract its syntactic role  (subject, object or other).
We use a  wikipedia dump parsed using  the Fanse Parser \cite{tratz2011fast}.
Subjects and objects are extracted based on {\it nsubj} and {\it dobj} relations in the dependency trees.
\cite{barzilay2008modeling} define two versions of the Entity Grid Model, one
using full coreference and a simpler method using only exact-string coreference; 
%Due to the hardness of running full coreference resolution over the size of training corpus we consider, 
Due to the difficulty of running full coreference resolution tens of millions of Wikipedia sentences,
we follow other researchers in using Barzilay and Lapata's simpler method \cite{feng2012extending,burstein2010using,barzilay2008modeling}.\footnote{Our implementation of the Entity Grid Model is built upon public available code at \url{https://github.com/karins/CoherenceFramework.}}

\paragraph{\bf Kendall's $\tau$}
Kendall's $\tau$ is computed based on the number
of inversions in the rankings as follows:
\begin{equation}
\tau=1-\frac{2 \text{\# of inversions}}{N\times (N-1)}
\end{equation} 
where $N$ denotes the number of sentences in the original document 
and inversions denote the number of interchanges of consecutive elements needed to reconstruct the original document. 
Kendall's $\tau$ can be efficiently computed by counting the number of 
intersections of lines when aligning the original document and the generated document. We refer the readers to \newcite{lapata2003probabilistic} for more details.

\paragraph{Derivation for Variation Inference}
For simplicity, we use 
$\mu_{post}$ and $\Sigma_{ approx}$ to denote $\mu^{\text{ approx}}({z_n})$ and
$\Sigma^{\text{ approx}}({z_n})$, 
$\mu_{true}$ and $\Sigma_{true}$ to denote $\mu^{\text{true}}({z_n})$ and
$\Sigma^{\text{true}}({z_n})$.
The
KL-divergence between the approximate distribution $q(z_n|z_{n-1},s_n,s_{n-1},...)$ and the true distribution $p(z_n|z_{n-1},s_{n-1},s_{n-2},...)$ in the variational inference is given by:
\begin{equation}
\begin{aligned}
&D_{KL}(q(z_n|z_{n-1},s_n,s_{n-1},...)||p(z_n|z_{n-1},s_{n-1},s_{n-2},...)\\
=&\frac{1}{2}(\text{tr}(\Sigma^{-1}_{true}\Sigma_{ approx})-k+\log\frac{\text{det}\Sigma_{true} }{\text{det}\Sigma_{ approx}}\\
+&(\mu_{true}-\mu_{ approx})^{-1}\Sigma^{-1}_{true}(\mu_{true}-\mu_{ approx}))
\end{aligned}
\label{Eq10}
\end{equation}
where $k$ denotes the dimensionality of the vector.
Since $z_n$ has already been sampled and thus known, 
$\mu_{ approx}$, $\Sigma_{ approx}$, $\mu_{true}$, $\Sigma_{true}$ 
and consequently 
Eq\ref{Eq10} can be readily computed. 
The gradient with respect to $\mu_{ approx}$, $\Sigma_{ approx}$, $\mu_{true}$, $\Sigma_{true}$  can be respectively computed, and the error is then backpropagated to the hierarchical neural models that are used to compute them. 
We refer the readers to \newcite{doersch2016tutorial} for more details about how a general VAE model can be trained.

Our generate models offer a powerful way to represent the latent
discourse structure in a complex embedding space, but one that is
hard to visualize.  To help understand what the model is doing, we 
examine some relevant examples, annotated with the  (log-likelihood)
coherence score from the{ \it MMI} generative model,
with the goal of seeing (qualitatively) the kinds of coherence
the model seems to be representing.
(The MMI can be viewed as the informational gain from conditioning the generation of the current sentence on its neighbors.)

\end{document}